\documentclass[sigconf,nonacm]{acmart}
\usepackage[ruled,linesnumbered]
{algorithm2e}
\usepackage{multirow}
\usepackage[normalem]{ulem}
\usepackage{url}
\usepackage{hyperref}
\useunder{\uline}{\ul}{}

\AtBeginDocument{%
  }

\setcopyright{acmlicensed}
\copyrightyear{2018}
\acmYear{2018}
\acmDOI{XXXXXXX.XXXXXXX}
\acmConference[Conference acronym 'XX]{Make sure to enter the correct
  conference title from your rights confirmation email}{June 03--05,
  2018}{Woodstock, NY}
\acmISBN{978-1-4503-XXXX-X/2018/06}

\begin{document}

%%
%% The "title" command has an optional parameter,
%% allowing the author to define a "short title" to be used in page headers.
\title{H3M-SSMoEs: Hypergraph-based Multimodal Learning with LLM Reasoning and Style-Structured Mixture of Experts}

%%
%% The "author" command and its associated commands are used to define
%% the authors and their affiliations.
%% Of note is the shared affiliation of the first two authors, and the
%% "authornote" and "authornotemark" commands
%% used to denote shared contribution to the research.

\author{Peilin Tan}
\affiliation{%
  \institution{University of California, San Diego}
  \city{La Jolla}
    \state{CA}
  \country{USA}}
\email{p9tan@ucsd.edu}

\author{Liang Xie}
\authornote{Corresponding author.}
\affiliation{%
  \institution{Wuhan University of Technology}
  \city{Wuhan, Hubei}
  \country{China}
}
\email{whutxl@hotmail.com}

\author{Churan Zhi}
\authornote{All three authors contributed equally to this research.}
\author{Dian Tu}
\authornotemark[2]
\author{Chuanqi Shi}
\authornotemark[2]
\affiliation{%
  \institution{University of California, San Diego}
  \city{La Jolla}
    \state{CA}
  \country{USA}
}
\email{chzhi@ucsd.edu}
\email{ditu@ucsd.edu}
\email{chs028@ucsd.edu}

%%
%% By default, the full list of authors will be used in the page
%% headers. Often, this list is too long, and will overlap
%% other information printed in the page headers. This command allows
%% the author to define a more concise list
%% of authors' names for this purpose.
% \renewcommand{\shortauthors}{Trovato et al.}

%%
%% The abstract is a short summary of the work to be presented in the
%% article.
\begin{abstract}
Stock movement prediction remains fundamentally challenging due to complex temporal dependencies, heterogeneous modalities, and dynamically evolving inter-stock relationships. Existing approaches often fail to unify structural, semantic, and regime-adaptive modeling within a scalable framework. This work introduces H3M-SSMoEs, a novel \textbf{H}ypergraph-based \textbf{M}ulti\textbf{M}odal architecture with LL\textbf{M} reasoning and \textbf{S}tyle-\textbf{S}tructured \textbf{M}ixture \textbf{o}f \textbf{E}xpert\textbf{s}, integrating three key innovations: (1) a Multi-Context Multimodal Hypergraph that hierarchically captures fine-grained spatiotemporal dynamics via a Local Context Hypergraph (LCH) and persistent inter-stock dependencies through a Global Context Hypergraph (GCH), employing shared cross-modal hyperedges and Jensen–Shannon Divergence weighting mechanism for adaptive relational learning and cross-modal alignment; (2) a LLM-enhanced reasoning module, which leverages a frozen large language model with lightweight adapters to semantically fuse and align quantitative and textual modalities, enriching representations with domain-specific financial knowledge; and (3) a Style-Structured Mixture of Experts (SSMoEs) that combines shared market experts and industry-specialized experts, each parameterized by learnable style vectors enabling regime-aware specialization under sparse activation. Extensive experiments on three major stock markets demonstrate that H3M-SSMoEs surpasses state-of-the-art methods in both superior predictive accuracy and investment performance, while exhibiting effective risk control. \textbf{Datasets, source code, and model weights are available at our GitHub repository: https://github.com/PeilinTime/H3M-SSMoEs}.
\end{abstract}

%%
%% The code below is generated by the tool at http://dl.acm.org/ccs.cfm.
%% Please copy and paste the code instead of the example below.
%%

% \ccsdesc[500]{Do Not Use This Code~Generate the Correct Terms for Your Paper}
% \ccsdesc[300]{Do Not Use This Code~Generate the Correct Terms for Your Paper}
% \ccsdesc{Do Not Use This Code~Generate the Correct Terms for Your Paper}
% \ccsdesc[100]{Do Not Use This Code~Generate the Correct Terms for Your Paper}

%%
%% Keywords. The author(s) should pick words that accurately describe
%% the work being presented. Separate the keywords with commas.
%% A "teaser" image appears between the author and affiliation
%% information and the body of the document, and typically spans the
%% page.

% \begin{teaserfigure}
%   \includegraphics[width=\textwidth]{sampleteaser}
%   \caption{Seattle Mariners at Spring Training, 2010.}
%   \Description{Enjoying the baseball game from the third-base
%   seats. Ichiro Suzuki preparing to bat.}
%   \label{fig:teaser}
% \end{teaserfigure}

% \received{20 February 2007}
% \received[revised]{12 March 2009}
% \received[accepted]{5 June 2009}

%%
%% This command processes the author and affiliation and title
%% information and builds the first part of the formatted document.
\maketitle

\section{Introduction}

Stock markets are fundamental to the global financial system, with accurate price prediction directly impacting capital allocation, portfolio optimization, and risk management. While the Efficient Market Hypothesis \cite{merello2019ensemble} suggests that prices reflect all available information, making future movements theoretically unpredictable, research has identified systematic inefficiencies—information asymmetry, behavioral biases, and market microstructure effects—that create potentially exploitable patterns for those capable of modeling and uncovering insights within complex market dynamics.

Predicting stock market movements remains an exceptionally challenging task due to a range of intertwined factors. Financial markets typically exhibit a low signal-to-noise ratio, where meaningful patterns are often obscured by random fluctuations. Their inherently non-stationary nature means that profitable patterns in one regime may fail as conditions change. 
Stock movements also involve complex interdependencies through sectoral correlations and momentum spillovers that evolve dynamically. 
In addition, prices react to influences operating across multiple timescales, while relevant information spans diverse forms, from structured numerical data to unstructured text.

Graph Neural Networks (GNNs) \cite{zhang2020deep} have emerged as a powerful framework for stock market prediction by modeling inter-stock relationships through industry affiliations and correlation, enabling information propagation across stocks to capture sectoral influences and spillover effects. However, conventional graph models are inherently limited to pairwise relationships. In contrast, real-world markets often exhibit complex group-wise correlations. Stocks within the same sector tend to move synchronously during sectoral shifts, and related industries experience collective movements under supply chain disruptions. These limitations motivate the adoption of hypergraphs \cite{agarwal2006higher, feng2019hypergraph}, where hyperedges can connect multiple nodes simultaneously, naturally encoding group relationships. Hypergraph representations preserve higher-order market structures and facilitate efficient computation by directly modeling group interactions rather than degenerating them into oversimplified binary relations.

Models that rely exclusively on numerical data exhibit fundamental epistemic constraints, as they are unable to anticipate phenomena absent from historical data. Corporate disclosures, regulatory shifts, and geopolitical events typically manifest as textual information prior to influencing market prices. The advent of Large Language Models (LLMs) \cite{chen2023chatgpt} has introduced new opportunities for processing textual data at scale. Equipped with extensive pre-trained knowledge of economics and finance, LLMs can assimilate dynamic news flows, thereby addressing informational gaps that traditional time series models are unable to bridge \cite{xie2023pixiu}.

Recent research has explored several strategies for integrating LLMs with quantitative models, including alignment method that maps time series into textual embeddings \cite{ding2023integrating}, and prompt-based approach \cite{li2024alphafin} that textualizes numerical data. Despite these advances, substantial challenges persist in achieving seamless LLM integration. Existing methods generally treat structural and textual information in isolation, thereby foregoing potential synergies. Furthermore, the inherent mismatch between the discrete, token-based processing of LLMs and the continuous nature of time series remains only partially resolved, underscoring the necessity for sophisticated multimodal fusion frameworks.

As models grow increasingly complex, computational efficiency becomes paramount. The Mixture of Experts (MoEs) framework \cite{jacobs1993adaptive, fedus2022switch, riquelme2021scaling} addresses this challenge by dynamically routing inputs to specialized expert networks, activating only relevant model subsets. This selective activation allows different experts to specialize in particular market conditions or sectors, preserving model capacity while maintaining manageable inference costs for practical deployment. 
However, integrating MoEs architectures with advanced financial modeling components remains nontrivial. Current approaches to combining hypergraph structures with transformer architectures often rely on simple feature fusion rather than joint reasoning mechanisms. Moreover, existing MoEs implementations typically fail to capture the hierarchical and multi-scale nature of market dynamics.
These limitations highlight the need for novel architectures capable of unifying relational modeling, modality alignment, textual understanding, and computational efficiency while remaining feasible for deployment in real trading environments.

To tackle these challenges, we propose a novel multi-modal architecture that synergistically integrates multi-context hypergraph modeling, LLM-enhanced semantic reasoning, and style–structure expert specialization. Our contributions are threefold:
\begin{itemize}

\item \textbf{Multi-Context Multimodal Hypergraph}: We introduce a hierarchical architecture consisting of a Local Context Hypergraph (LCH) that captures fine-grained spatiotemporal dynamics at the instance level, and a Global Context Hypergraph (GCH) that models persistent structural dependencies across stocks. 
Both components utilize shared hyperedges that jointly connect nodes from quantitative and textual modalities, enabling direct interaction between market signals and news narratives. Through hypergraph convolutions, these shared connections facilitate mutual representation enhancement and cross-modal feature learning, achieving deep, integrated multi-modal understanding beyond simple fusion.

\item \textbf{LLM-Enhanced Reasoning}: We incorporate a frozen Large Language Model (Llama-3.2-1B) to bridge the semantic gap between textual and numerical information. Leveraging its pre-trained financial knowledge, the LLM enriches multi-modal representations while preserving efficiency via parameter freezing and lightweight adapter layers.

\item \textbf{Style-Structured Mixture of Experts (SSMoEs)}: We introduce a MoEs module with learnable style parameters that enables adaptive specialization across different market states and industry conditions via sparse activation. This design maintains computational efficiency while preserving high model capacity and complements the hypergraph for robust regime-aware representations.

\end{itemize}
Extensive experiments on the DJIA, NASDAQ 100, and S\&P 100 indices demonstrate our method's state-of-the-art performance, achieving the highest risk-adjusted returns with Sharpe ratios of 1.585, 2.100, and 1.351, and Calmar ratios of 3.377, 4.380, and 2.075, respectively, while maintaining the lowest maximum drawdowns (14.81\%, 16.17\%, and 14.27\%).

\section{Related Work}

\subsection{Graph \& Hypergraph for Stock Relations}

Early stock prediction methods primarily relied on statistical models \cite{wang1996stock,zivot2006vector}, which assume linear dependencies and thus struggle to capture the complex dynamics of financial markets. Subsequent machine learning approaches \cite{ballings2015evaluating} enhanced non-linear modeling capabilities but often treated stocks independently, overlooking inter-stock dependencies. This limitation motivated the adoption of graph-based models to represent the inherent relational structure among stocks.

Recognizing that stock movements are highly interconnected, researchers began employing Graph Neural Networks (GNNs) to model inter-stock relationships \cite{chen2018incorporating}. Early studies constructed graphs using predefined relationships such as common shareholders, industry sectors, or supply chains. More advanced models, including RSR \cite{feng2019temporal}, which integrates LSTM with graph convolutions, HATS \cite{kim2019hats}, which introduces multi-relational attention mechanisms, and FinGAT \cite{hsu2021fingat}, which applies dynamic attention to quantify stock interactions, have demonstrated improved performance. Recent approaches have shifted from static, predefined structures to dynamically learned relationships that capture latent dependencies between stocks \cite{sawhney2021stock}.

Beyond pairwise relations, the recognition of group-wise interactions has spurred the development of hypergraph-based models. STHGCN \cite{sawhney2020spatiotemporal} jointly models the temporal evolution of stock prices and their industry-level associations, effectively capturing higher-order dependencies. More recently, CI-STHPAN \cite{xia2024ci} introduced a pre-training framework on stock time series followed by fine-tuning for quantitative stock selection, leveraging self-supervised learning to extract robust spatio-temporal representations.

\begin{figure*}[t]
\centering
\includegraphics[width=1\textwidth]{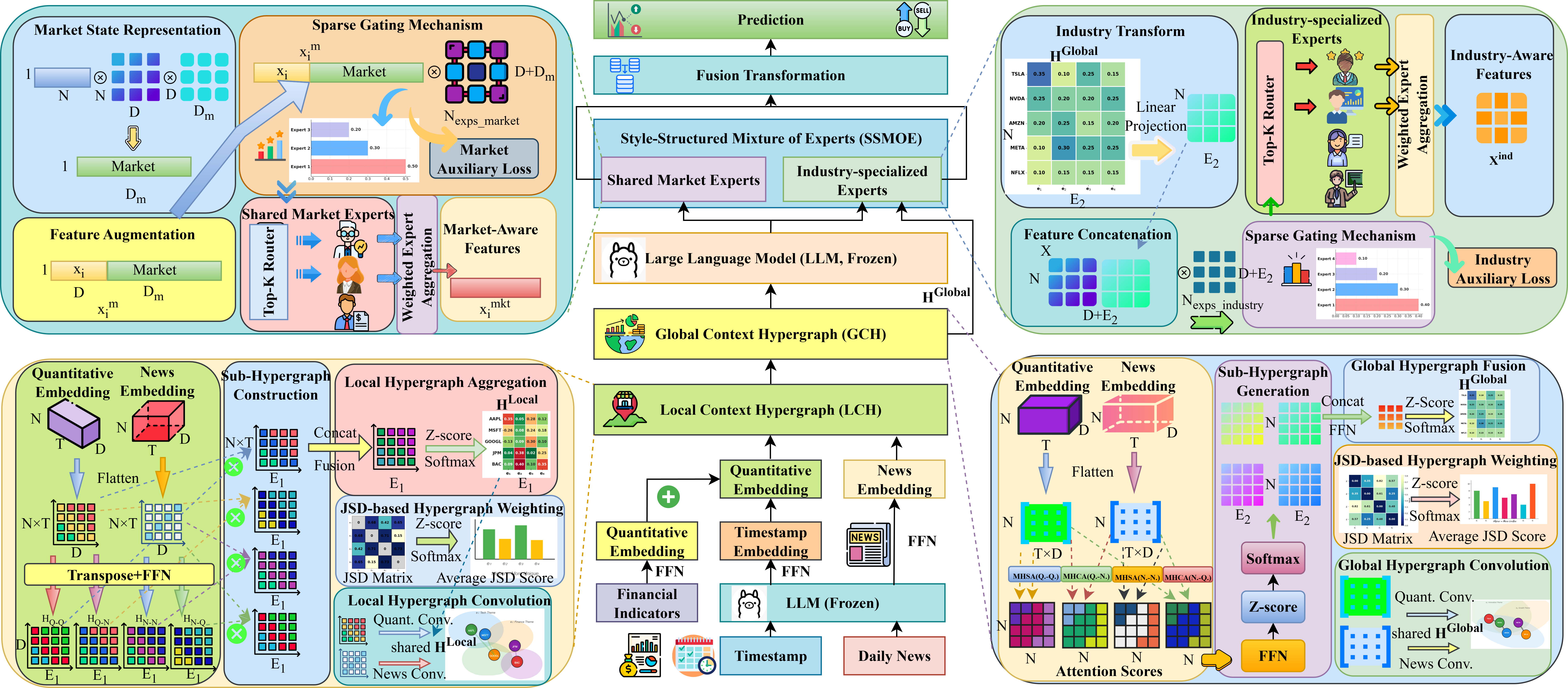}
\caption{Overview of the H3M-SSMoEs. The framework comprises:
(1) Feature embedding using a frozen LLM for textual data;
(2) Multi-Context Multimodal Hypergraph processing, including the Local Context Hypergraph (LCH) for capturing instance-level dependencies and the Global Context Hypergraph (GCH) for modeling inter-stock relationships;
(3) LLM-enhanced multimodal reasoning for deeper semantic integration; and
(4) Style-Structured Mixture of Experts (SSMoEs), which combines shared market experts and industry-specialized experts for adaptive, style-aware prediction.}
\label{Overview}
\end{figure*}

\subsection{LLM \& Foundation Models in Finance}

Deep learning has revolutionized stock market prediction through diverse neural architectures. Recurrent Neural Networks (RNNs) \cite{cho2014learning}, particularly LSTM \cite{hochreiter1997long, moghar2020stock} and GRU \cite{gupta2022hisa} variants, have been widely utilized for their ability to capture sequential dependencies and temporal dynamics. The DA-RNN \cite{qin2017dual} introduced LSTMs to adaptively extract relevant features, while the State Frequency Memory (SFM) network \cite{zhang2017stock} decomposed hidden states into multiple frequency components to enhance representational diversity. Moreover, transformer-based architectures have achieved superior performance by effectively modeling complex temporal and cross-asset dependencies. Stockformer \cite{ma2025stockformer} integrates wavelet decomposition with dual-frequency spatio-temporal encoders and a fusion attention mechanism to capture both high- and low-frequency financial dynamics.

Recently, Large Language Models (LLMs) have emerged as a new paradigm in financial modeling. BloombergGPT \cite{wu2024bloomberggpt} represents a seminal effort, comprising 50 billion parameters trained on a hybrid corpus of financial and general-domain texts. FinGPT \cite{yang2023fingptopensourcefinanciallarge} and DISC-FinLLM \cite{chen2023discfinllmchinesefinanciallarge} introduced instruction fine-tuning and low-rank adaptation, to improve task-specific performance. LLMs also have been leveraged under various paradigms: news-driven approaches utilize sentiment and contextual analysis of market narratives, while reasoning-driven agents like FinMEM \cite{yu2023finmemperformanceenhancedllmtrading} and FinAgent \cite{zhang2024multimodalfoundationagentfinancial} incorporate multimodal data sources, such as earnings calls, regulatory filings, and social media, to augment predictive accuracy and interpretability.

Time Series Foundation Models (TSFMs) have further extended the foundation model paradigm to temporal data. Models such as TimesFM \cite{das2024decoderonlyfoundationmodeltimeseries}, Lag-Llama \cite{rasul2024lagllamafoundationmodelsprobabilistic}, and Chronos \cite{ansari2024chronoslearninglanguagetime} exhibit strong generalization across diverse time series domains by pre-training on large-scale temporal corpora. Financially specialized models, including Kronos \cite{shi2025kronosfoundationmodellanguage}, address a key limitation of general-purpose TSFMs by focusing exclusively on financial datasets, thereby improving domain relevance. To mitigate computational costs and improve scalability, Mixture of Experts (MoE) architectures have emerged, with Time-MoE \cite{shi2025timemoebillionscaletimeseries} scales TSFMs to billions of parameters while achieving 20–24\% performance gains over dense models under equivalent computational budgets.

\subsection{Multimodal Financial Forecasting}

The integration of numerical and textual  data in time series forecasting has progressed from rudimentary keyword-based approaches to advanced architectures that leverage the full representational capacity of large language models. Time-LLM \cite{jin2024timellmtimeseriesforecasting} reprograms time series into text-like representations compatible with LLM embedding spaces, facilitating multimodal interaction. Similarly, ChatTime \cite{wang2024chattimeunifiedmultimodaltime} conceptualizes time series as a foreign language, converting continuous numerical sequences into discrete tokens. TGForecaster \cite{xu2025interventionawareforecastingbreakinghistorical} employs PatchTST encoders \cite{nie2023timeseriesworth64} for temporal data while incorporating pre-trained text models to process news, achieving efficient cross-modal fusion.

For financial forecasting and stock prediction, MoE architectures have not yet been systematically explored, but they present substantial potential for advancing model adaptability and scalability. Given the inherently heterogeneous nature of financial markets characterized by regime shifts, sectoral dependencies, and varying volatility structures, MoE provides a natural framework for modular specialization. Each expert can be tailored to capture specific temporal patterns, market trends, industry behaviors and personalized sentiment, while sparse gating mechanisms ensure computational efficiency. Despite these advantages, the integration of MoE with multimodal architectures remains an unexplored frontier. Existing approaches typically treat structural representation, temporal modeling, and textual understanding in isolation. A promising research direction lies in developing unified frameworks that jointly incorporate hypergraph-informed structural priors, LLM-based semantic reasoning, and specialized MoE processing, balancing representational richness with efficiency.

\section{Methodology}

We formulate the $d$-day-ahead stock movement prediction as a binary classification problem, aiming to forecast whether the closing price of each constituent stock within the market index will rise after $d$ trading days.
THe H3M-SSMoEs leverages three complementary modalities:
(1) Historical quantitative features, extracted over a $T$-day lookback window;
(2) Daily financial news, encoded using the frozen Llama-3.2-1B large language model to capture semantic context; 
and (3) Timestamp embedding, which encode explicit temporal information using the same frozen LLM.
Detailed definition is provided in Appendix \ref{Problem Definition}.

Figure~\ref{Overview} illustrates the overall architecture of the H3M-SSMoEs, which integrates three key components: Multi-Context Multimodal Hypergraph modeling, LLM-driven semantic enhancement with multimodal reasoning, and the Style-Structured Mixture of Experts (SSMoEs) to enable adaptive and context-aware stock movement prediction.

\subsection{Feature Embedding}

To facilitate cross-modal learning, all heterogeneous input modalities are projected into a unified latent space of dimension $D$. These projections enable the alignment of diverse data sources—quantitative indicators, textual news, and temporal information—within a common representational framework.

\textbf{Modality-Specific Projection.}  
Each modality $m$ (quantitative, news, or timestamps) is transformed through a modality-specific feed-forward network that maps the original input space to the shared latent dimension:
\begin{equation}
\mathbf{h}_{n,t}^{(m)} = \text{FFN}^{(m)}_{proj}(\mathbf{x}_{n,t}^{(m)}) \in \mathbb{R}^{D},
\end{equation}
where $\mathbf{x}_{n,t}^{(m)}$ denotes the input features of modality $m$ for stock $n$ at time $t$. Specifically, $\mathbf{x}_{n,t}^{\text{quant}} \in \mathbb{R}^{F}$ corresponds to $F$ financial indicators, $\mathbf{x}_{n,t}^{\text{news}} \in \mathbb{R}^{D_{\text{news}}}$ represents pre-computed LLM embeddings of news, and $\mathbf{x}_t^{\text{time}} \in \mathbb{R}^{D_{\text{time}}}$ encodes timestamps via LLM-processed date representations shared across all stocks.

\textbf{Temporal Encoding Integration.}
To introduce temporal awareness into quantitative features and strengthen their alignment with daily news embeddings, we incorporate positional encodings as follows:
\begin{equation}
\mathbf{z}_{n,t}^{(m)} = \begin{cases}
\mathbf{h}_{n,t}^{\text{quant}} + \mathbf{h}_t^{\text{time}}, & \text{if } m = \text{quantitative} \\
\mathbf{h}_{n,t}^{\text{news}}, & \text{if } m = \text{news}
\end{cases}.
\end{equation}
This formulation explicitly injects temporal context into quantitative representations, enhancing their ability to capture time-dependent market dynamics. In contrast, news embeddings inherently encode temporal semantics through their linguistic content. The resulting representations form two parallel embedding streams, $\mathbf{Z}^{\text{quant}}, \mathbf{Z}^{\text{news}} \in \mathbb{R}^{N \times T \times D}$, which serve as inputs for cross-modal alignment within subsequent hypergraph-based layers.

\subsection{Multi-Context Multimodal Hypergraph}

Traditional graph-based approaches are inherently constrained by predefined static pairwise relationships, which limits their ability to capture dynamic, collective market behaviors where entire sectors often move synchronously. To address this limitation, we introduce a multi-context multimodal hypergraph framework that hierarchically models both local and global interactions through higher-order relationships. Our architecture leverages shared hypergraph to unify intra-modal and cross-modal interactions, facilitating comprehensive information exchange between quantitative signals and semantic news semantics. This design effectively captures the bidirectional interplay between the two modalities—how news drives market movements and how market fluctuations, in turn, generate new narratives. By modeling these intertwined dependencies, the proposed framework achieves synergistic multimodal integration and cross-modal alignment, surpassing conventional fusion approaches.

\subsubsection{Local Context Hypergraph (LCH)}

The Local Context Hypergraph (LCH) is designed to model the intricate spatiotemporal dependencies inherent in financial markets, where individual stock movements are jointly influenced by immediate behavioral patterns and market narratives. Unlike conventional methods relying on fixed temporal windows or predefined, static correlations, the LCH flexibly discovers dynamic group relationships of stock–time quantitative and textual instances that exhibit coordinated behaviors by representing each stock at each timestamp as a distinct node within a hypergraph. This formulation preserves fine-grained temporal resolution while leveraging hyperedges to model group-wise dependencies that evolve over time.

To achieve unified processing across temporal and spatial dimensions, we flatten the stock and time dimensions to obtain $\mathbf{Z}^{(m)}_{flat} \in \mathbb{R}^{N \cdot T \times D}$ for each modality $m \in \{\text{quantitative}, \text{news}\}$, where each row represents a unique stock–time instance. The core innovation lies in constructing multimodal sub-hypergraphs that capture distinct types of dependencies: temporal correlations within numerical indicators, semantic coherence across news narratives, and the bidirectional interplay between quantitative and textual modalities. This design explicitly acknowledges that market behaviors are governed by fundamentally heterogeneous forms of relationships and modalities.
For each modality pair $(m_i, m_j)$ (quantitative or news), a specialized sub-hypergraph is learned via adaptive projection:
\begin{equation}
    \mathbf{H}^{(m_i, m_j)}_{local} = \mathbf{Z}^{(m_i)}_{flat} \cdot \text{FFN}_{local}^{(m_i, m_j)}((\mathbf{Z}^{(m_i)}_{flat})^T) \in \mathbb{R}^{(N\cdot T) \times E_1},
\end{equation}
where $\text{FFN}_{local}^{(m_i, m_j)}$ is a modality-pair-specific network that learns to identify $E_1$ latent hyperedges, each representing a set of stock–time instances exhibiting coordinated behaviors. This formulation yields four distinct sub-hypergraphs corresponding to intra- and inter-modal relationships intertwined with critical market dynamics:
\begin{itemize}
    \item \textbf{Quantitative–Quantitative Dynamics}: Captures temporal momentum and volatility clustering within numerical market indicators;
    
    \item \textbf{News–News Coherence}: Models semantic coherence and the propagation of narratives across the news landscape;
    
    \item \textbf{Quantitative–News Alignment}: Aligns market reactions with contemporaneous news events. This cross-modal sub-hypergraph learns how current price patterns co-occur with specific news narratives, contextualizing stock movements within the textual information;
    
    \item \textbf{News–Quantitative Anticipation}: Represents the inverse relationship—how news content anticipates market movements. This component captures predictive cues embedded in texts that may not yet manifest in price dynamics.
\end{itemize}
This symmetric design explicitly differentiates between the market’s reactive and anticipatory responses to news information.

Although these sub-hypergraphs encode complementary aspects of market behavior, their relative importance fluctuates with changing market regimes. Rather than treating all relationships uniformly, the LCH employs an adaptive multi-hypergraph fusion mechanism:
\begin{equation}
    \mathbf{H}_{local}' = \text{FFN}_{local}^{\text{fusion}}([\mathbf{H}^{(m_i,m_j)}_{local}]_{\text{all pairs}}) \in \mathbb{R}^{(N \cdot T) \times E_1}.
\end{equation}
This fusion network dynamically weights and integrates the intra- and cross-modal sub-hypergraphs in accordance with the prevailing market context. Subsequently, z-score normalization is applied to each element within the hyperedges, followed by a column-wise softmax operation to ensure that each hyperedge constitutes a probability distribution over nodes, resulting in the unified incidence matrix $\mathbf{H}_{local}$.

However, not all hyperedges contribute equally to modeling market structure. Some effectively encode distinctive and insightful dependencies, whereas others capture redundant or noisy patterns. To highlight the most informative structures, we introduce an information-theoretic hyperedge weighting scheme based on the Jensen–Shannon Divergence (JSD). For each pair of hyperedges $i$ and $j$:
\begin{equation}
    \text{JSD}(i,j) = \frac{1}{2}\big[\text{KL}(\mathbf{h}_i \,||\, \mathbf{m}_{ij}) + \text{KL}(\mathbf{h}_j \,||\, \mathbf{m}_{ij})\big],
\end{equation}
where $\mathbf{h}_i$ denotes the node distribution of the $i$-th hyperedge, $\mathbf{m}_{ij} = \frac{1}{2}(\mathbf{h}_i + \mathbf{h}_j)$ is their mean distribution, and KL is the Kullback–Leibler Divergence. Hyperedges with higher average JSD scores capture more unique relational structures and thus receive larger weights in the diagonal matrix $\mathbf{W_1} \in \mathbb{R}^{E_1 \times E_1}$. This adaptive weighting encourages the model to emphasize structurally informative hyperedges while suppressing redundant ones.

Finally, information is propagated through the weighted hypergraph structure to model higher-order interactions among stock–time instances. For each modality, hypergraph convolution is performed using the shared local hypergraph $\mathbf{H}^{local}$:
\begin{equation}
    \mathbf{Z'}^{(m)}_{LCH} = \sigma(\mathbf{H}_{local} \mathbf{W_1} \mathbf{H}_{local}^T \mathbf{Z}^{(m)}_{flat} \boldsymbol{\Theta}^{(m)}_{local}),
\end{equation}
where $\boldsymbol{\Theta}^{(m)}_{local} \in \mathbb{R}^{D \times D}$ is a learnable modality-specific transformation matrix, and $\sigma$ denotes a nonlinear activation. This operation facilitates high-order cross-temporal, and cross-modal interactions, enabling implicit alignment between modalities. Consequently, the model learns temporal lead–lag relationships and cross-stock spillover effects, uncovering latent connections between quantitative market dynamics and textual narratives. 
The resulting features $\mathbf{Z'}^{(m)}_{LCH}$ are then reshaped to $\mathbb{R}^{N \times T \times D}$ to yield $\mathbf{Z}^{(m)}_{LCH}$, embedding sophisticated dependencies.

\subsubsection{Global Context Hypergraph (GCH)}

While the Local Context Hypergraph (LCH) captures fine-grained spatiotemporal patterns, the Global Context Hypergraph (GCH) models persistent structural relationships—such as sector affiliations, supply chain dependencies, and competitive dynamics—that span the entire temporal horizon. This global perspective facilitates the identification of stable sectoral structures and long-term dependencies through higher-order group interactions.

The GCH adopts an architecture analogous to that of the LCH but operates at the stock level rather than the instance level, thereby modeling persistent inter-stock relationships across time. The features are first flattened into stock-level representations of shape $\mathbb{R}^{N \times (T \cdot D)}$. Subsequently, multi-head self- and cross-attention mechanisms, sub-hypergraph construction, adaptive fusion, and the derivation of the global hypergraph incidence matrix $\mathbf{H}_{global}$ are performed, followed by the JSD-based weighting mechanism and hypergraph convolution to derive enhanced global representations for both modalities. These representations are then reshaped back to $\mathbb{R}^{N \times T \times D}$, yielding $\mathbf{Z}^{(m)}_{GCH}$. Comprehensive formulations of these processes are provided in Appendix \ref{Global Context Hypergraph (GCH)}.

The GCH captures industry-wide trends, sectoral correlations, and market-wide sentiment flows, thereby complementing the micro-level temporal patterns learned by the LCH. Working in concert, these two hypergraphs yield a multi-context representation: the LCH effectively models short-term responses—such as how individual stocks react to technical indicators or news—while the GCH contextualizes these reactions within overarching market dynamics. Collectively, this hypergraph architecture forms a robust foundation for comprehensive market understanding, enabling rich multi-modal and multi-scale representations.

\subsection{LLM for Semantic Enhancement and Multimodal Reasoning}

Following hypergraph processing, we incorporate a frozen Large Language Model (LLM) to enrich semantic representations and facilitate advanced multimodal reasoning across the aligned numerical and textual modalities. Specifically, the quantitative and news embeddings $\mathbf{Z}^{\text{quant}}_{GCH}, \mathbf{Z}^{\text{news}}_{GCH} \in \mathbb{R}^{N \times T \times D}$ produced by the GCH are further refined through an LLM-based reasoning layer.

We employ the frozen Llama-3.2-1B model, chosen for its favorable trade-off between semantic reasoning capacity and computational efficiency. Freezing the LLM parameters preserves its extensive linguistic and financial domain-specific knowledge acquired during pre-training, while lightweight adapter layers are utilized to perform modality alignment and feature fusion without imposing significant training costs.

First, the quantitative and news features are concatenated along the feature dimension:
\begin{equation}
\mathbf{Z}_{\text{cat}} = [\mathbf{Z}^{\text{quant}}_{GCH}, \mathbf{Z}^{\text{news}}_{GCH}] \in \mathbb{R}^{N \times T \times 2D}.
\end{equation}
The concatenated representations are then projected into the LLM input space via a multimodal fusion network:
\begin{equation}
\mathbf{Z}_{\text{fused}} = \text{FFN}_{\text{fusion}}(\mathbf{Z}_{\text{cat}}) \in \mathbb{R}^{N \times T \times D_{\text{LLM}}},
\end{equation}
where $D_{\text{LLM}} = 2048$ denotes the hidden dimension of the Llama-3.2-1B model. The fused embeddings are subsequently processed by the frozen LLM to yield high-level semantic representations:
\begin{equation}
\mathbf{Z}_{\text{LLM}} = \text{LLM}(\mathbf{Z}_{\text{fused}}) \in \mathbb{R}^{N \times T \times D_{\text{LLM}}}.
\end{equation}
This design enables the framework to exploit the LLM’s pre-trained understanding of finance, market structures, and contextual semantics while maintaining computational efficiency through parameter freezing. The LLM serves as a semantic reasoning engine that enhances multimodal feature representations with deep linguistic and financial knowledge, thereby augmenting the model’s capacity for nuanced and context-aware market prediction.

\subsection{Style-Structured Mixture of Experts (SSMoEs)}

Stock markets exhibit heterogeneous behaviors across multiple scales—from global sentiment shifts to sector-specific momentum. We propose a Style-Structured Mixture of Experts (SSMoEs) architecture comprising two complementary expert pools: Shared Market Experts, which model overarching market regimes, and Industry-specialized Experts, which capture sector-level dynamics. Leveraging sparse activation, only the most relevant experts are dynamically selected based on market context and industry characteristics, ensuring high model expressiveness with efficient computation.

A central innovation of this module lies in the parametric style space: each expert incorporates learnable style parameters that enable distinct predictive strategies. 
During training, these parameters naturally differentiate across experts—yielding diverse strategic orientations such as bullish versus bearish or conservative versus aggressive. This diversity fosters a broad spectrum of adaptive trading perspectives among the experts.

\subsubsection{Shared Market Experts}

Individual stock dynamics are strongly conditioned by the broader market environment. Distinct market regimes, such as bullish, bearish, or high-volatility phases, exhibit systematic patterns that collectively shape asset behavior. The Shared Market Experts module is designed to infer the prevailing market regime and to adapt its specialization accordingly.

To capture the global market state, we first flatten $\mathbf{Z}_{LLM} \in \mathbb{R}^{N \times T \times D_{LLM}}$ into $\mathbb{R}^{N \times T \cdot D_{LLM}}$, yielding $\mathbf{Z}_{flat}$. This representation is then projected into a lower-dimensional space $\mathbb{R}^{N \times D}$ to reduce dimensionality and computational overhead. Next, we derive the market state by aggregating information across all stocks:
\begin{equation}
\mathbf{m} = W_l\mathbf{Z}_{flat}W_r \in \mathbb{R}^{D_m},
\end{equation}
where $W_r \in \mathbb{R}^{D \times D_m}$ projects features into a market-representative subspace, and $W_l \in \mathbb{R}^{1 \times N}$ performs cross-stock aggregation. These projections allow the model to learn both which features are most informative for market-state inference and how to optimally aggregate asset-level information.

The resulting market state is then concatenated with individual stock features to construct an augmented representation:
\begin{equation}
\mathbf{z}^{mkt}_i = [\mathbf{z}_i^{flat}, \mathbf{m}] \in \mathbb{R}^{D + D_m},
\end{equation}
ensuring that each stock’s routing decision reflects both local characteristics and global market context.

Each Shared Market Expert $j$ is parameterized by a learnable style vector $\mathbf{s}_j^{mkt} \in \mathbb{R}^{D_s}$, which defines its regime-specific specialization:
\begin{equation}
\text{Expert}_j^{mkt}(\mathbf{z}_i^{flat}) = \text{FFN}_j^{mkt}([\mathbf{z}_i^{flat}, \mathbf{s}_j^{mkt}]),
\end{equation}
Through training, these style vectors evolve into distinct market archetypes—such as bullish, bearish, or neutral—enabling the ensemble to capture a diverse and adaptive set of prediction and trading behaviors across varying market conditions.

\subsubsection{Industry-specialized Experts}

Beyond market-level influences, sector-specific forces often drive synchronized movements within industries, shaped by shared fundamentals or supply-chain dependencies. The Industry-specialized Experts complements the market view by modeling these intra-sector dynamics through experts specialized in distinct industry behaviors.

This module leverages the higher-order sectoral relationships $\mathbf{H}_{global}$ learned by the Global Context Hypergraph (GCH) to guide industry-aware routing:
\begin{equation}
\mathbf{I} = \text{FFN}_{\text{ind}}(\mathbf{H}_{global}) \in \mathbb{R}^{N \times E_2},
\end{equation}
The transformation extracts industry embeddings that capture latent cross-sector dependencies and evolving sectoral relationships. Each stock’s representation is then augmented as:
\begin{equation}
\mathbf{z}^{ind}_i = [\mathbf{z}_i^{flat}, \mathbf{I}_i] \in \mathbb{R}^{D+E_2},
\end{equation}
where $\mathbf{I}_i$ is the $i$-th row of $\mathbf{I}$, representing the industry-level embedding associated with stock $i$. This design provides the routing mechanism with both asset-specific and industrial context. For instance, a semiconductor stock might be routed to experts specializing simultaneously in technology momentum, global supply-chain shifts, and cyclical manufacturing patterns.

Each Industry-specialized Expert $k$ also maintains a learnable style vector $\mathbf{s}_k^{ind} \in \mathbb{R}^{D_s}$:
\begin{equation}
\text{Expert}_k^{ind}(\mathbf{z}_i) = \text{FFN}_k^{ind}([\mathbf{z}_i^{flat}, \mathbf{s}_k^{ind}]).
\end{equation}
These style vectors encourage differentiation among experts, promoting the emergence of specialized sectoral archetypes—some focusing on defensive industries, others on growth-oriented technology or cyclical manufacturing sectors.

Finally, both expert modules employ sparse gating with top-K selection to aggregate outputs from the most relevant experts in their respective pools, yielding $\mathbf{h}^{mkt}_i, \mathbf{h}^{ind}_i$ for each stock $i$, respectively. Detailed formulations of the routing and weighted aggregation processes are provided in Appendix \ref{Expert Routing and Aggregation}.

\subsubsection{Expert Pool Aggregation Layer}

The SSMoEs module integrates complementary insights from both global market and industrial perspectives via an flexible aggregation mechanism. This layer coordinates expert selection and aggregation across both pools, producing final predictions that adapt to multi-scale market structures.

Specifically, the final integration stage adaptively combines outputs from both expert pools through a learnable nonlinear fusion:
\begin{equation}
\mathbf{z}_{i} = \sigma(\mathbf{W}_{\text{mkt}}\mathbf{h}^{mkt}_i + \mathbf{W}_{\text{ind}}\mathbf{h}^{ind}_i),
\end{equation}
where $W_{\text{mkt}}, W_{\text{ind}} \in \mathbb{R}^{d \times d}$ are learnable weights controlling the relative influence of market and industry signals, and $\sigma(\cdot)$ denotes a nonlinear activation. By integrating broad and granular insights, the SSMoEs captures multi-scale dependencies, yielding representations that reflect each stock’s unique position within the evolving market ecosystem.

\subsection{Loss Function}

The fused representation $\mathbf{z}_i$ is finally passed through a FFN followed by softmax to generate the binary classification probabilities $\mathbf{\hat{y}}_i = [\hat{y}_{i,0}, \hat{y}_{i,1}]$ for each stock $i$, indicating the likelihood of an upward price movement $d$ days ahead.

To optimize the model, we employ a composite loss function that combines the classification objective with two auxiliary losses for balanced expert utilization. The classification component adopts the cross-entropy loss:
\begin{equation}
    \mathcal{L}_{\text{cls}} = -\frac{1}{N} \sum_{i=1}^{N} \left[ y_i \log \hat{y}_i + (1 - y_i) \log (1 - \hat{y}_i) \right],
\end{equation}
where $y_i \in \{0, 1\}$ denotes the ground-truth label for the $i$-th stock.

To encourage balanced expert utilization within both expert pools, we incorporate sequence-wise auxiliary losses \cite{liu2024deepseek}. For each module, the auxiliary loss is formulated as:
\begin{equation}
    \mathcal{L}_{\text{aux}}^{e} = \sum_{i=1}^{N_e} f_i P_i, \quad e \in \{\text{market}, \text{industry}\}
\end{equation}
where $f_i$ represents the fraction of stocks routed to expert $i$, $P_i$ denotes the average routing probability assigned to expert $i$, and $N_e$ is the total number of experts. This term promotes uniform expert utilization while preserving specialization among experts.

The overall loss function combines all components as:
\begin{equation}
    \mathcal{L} = \mathcal{L}_{\text{cls}} + \alpha  \mathcal{L}_{\text{aux}}^{\text{market}} + \beta  \mathcal{L}_{\text{aux}}^{\text{industry}},
\end{equation}
where $\alpha$ and $\beta$ are the  balance factors.

\section{Experiment}

\subsection{Features}

\subsubsection{Quantitative Data} 
We obtained historical stock data from Yahoo Finance \footnote{\url{https://ranaroussi.github.io/yfinance/}}, comprising five basic attributes: close, high, low, open, and volume. To enrich the features, we utilized Qlib \cite{yang2020qlibaiorientedquantitativeinvestment} to compute the Alpha158 and Alpha360 technical indicators. After removing features containing missing values, these indicators were integrated with the basic attributes to construct an enriched dataset. To prevent information leakage and promote stable model training, z-score normalization was applied independently within each data split.

\subsubsection{News Data} 
To complement the quantitative features with textual information that captures market sentiment and analytical insights, we employed the Finrobot \cite{yang2024finrobotopensourceaiagent} to generate daily news for each stock. The agent synthesizes multiple data sources, including recent financial news, company fundamentals, and market dynamics, to produce structured textual content. This approach ensures consistent and high-quality textual data across all stocks and time periods, effectively addressing the issue of incomplete or missing news coverage that often affects smaller firms or periods of low market activity. By integrating these rich textual features with quantitative data, we provide the model with comprehensive multimodal inputs that enhance its ability to predict stock movements accurately.

\subsection{Datasets}
We evaluated our method on three major stock indices: DJIA, NASDAQ 100, and S\&P 100, using data from January 1, 2020 to August 31, 2025. The dataset was split 7:1:2 into training, validation (for hyperparameter tuning), and testing (for final evaluation). See Table \ref{tab:datasets} for detailed statistics.

\begin{table}[tbh!]
\caption{Statistics of Datasets}
\label{tab:datasets}
\begin{center}
    % \begin{small}
    % \setlength{\tabcolsep}{1mm}
\begin{tabular}{ccccc}
\toprule
\textbf{Dataset} & \textbf{\# Stocks} & \textbf{\# Training} & \textbf{\# Val.} & \textbf{\# Test} \\ \midrule
\textbf{DJIA}       & 30  & 996 & 142 & 285 \\ 
\textbf{NASDAQ 100} & 91  & 996 & 142 & 285 \\ 
\textbf{S\&P 100}    & 99  & 996 & 142 & 285 \\ \bottomrule
\end{tabular}
% \end{small}
\end{center}
\end{table}

\subsection{Baselines}
We evaluate H3M-SSMoEs against 15 baselines, spanning 4 categories:
\begin{itemize}
    \item \textbf{Stock Prediction Models} (6): SFM \cite{zhang2017stock}, Adv-ALSTM \cite{feng2018enhancing}, DTML \cite{attention1}, ESTIMATE \cite{huynh2023efficient}, StockMixer \cite{fan2024stockmixer}, MASTER \cite{li2024master};
    \item \textbf{Time Series Models} (3): DLinear \cite{zeng2023transformers}, iTransformer \cite{liu2023itransformer}, TimeMixer \cite{wang2024timemixer};
    \item \textbf{Graph Models} (3): GCN \cite{kipf2016semi}, GraphSAGE \cite{hamilton2017inductive}, GAT \cite{velivckovic2017graph};
    \item \textbf{Time Series LLM \& Foundation Model} (3): GPT4TS \cite{zhou2023fitsallpowergeneraltime},
    aLLM4TS \cite{bian2024multipatchpredictionadaptingllms},
    Time-LLM \cite{jin2024timellmtimeseriesforecasting}.
\end{itemize}
Detailed experiment settings of our model are presented in Appendix \ref{Experiment Settings}, and descriptions of all baselines are provided in Appendix \ref{Baseline Descriptions}.

\subsection{Evaluation Metrics}

Model performance is evaluated using both portfolio backtesting and classification-based metrics to comprehensively assess investment returns and predictive accuracy.

\begin{itemize}
    \item \textbf{Backtesting}: Annual Return (AR), Sharpe Ratio (SR, applying a 2\% risk-free rate), Calmar Ratio (CR), and Maximum Drawdown (MDD) are employed to evaluate the profitability and risk  of the model within simulated investment scenarios.
    
    \item \textbf{Prediction}: Accuracy (ACC) and Precision (PRE) are used to measure the quality of the model’s classification performance, where Precision denotes the proportion of stocks predicted and purchased as ``rising'' that actually increased in closing price during the holding period.
\end{itemize}
Detailed definitions and formulations of evaluation metrics are provided in Appendix \ref{Metric Definitions}.

\begin{table}[htb]
\caption{Backtesting \& Prediction Results on DJIA. The best results are in bold and the second-best results are underlined.}
  \label{tab:Prediction & Backtesting Results on DJIA}
\begin{tabular}{ccccccc}
\toprule
\multirow{2}{*}{\textbf{Model}} & \multicolumn{2}{c}{\textbf{Prediction}}  & \multicolumn{4}{c}{\textbf{Backtesting}}                                   \\ \cmidrule(lr){2-3} \cmidrule(lr){4-7}
                   & \textbf{ACC}            & \textbf{PRE}            & \textbf{AR}             & \textbf{SR}             & \textbf{CR}             & \textbf{MDD}            \\
\midrule
\textbf{DLin.}            & \textbf{57.50} & 58.08          & 15.92          & 0.889          & 0.963          & 16.53          \\

\textbf{iTrans.}       & 55.74          & 58.76          & 28.33          & 1.252          & 1.329          & 21.32          \\

\textbf{TimeM.}          & 57.27          & 55.31          & 5.16           & 0.256          & 0.280          & 18.43          \\

\textbf{GCN}                & 52.77          & 58.70          & 22.49          & 1.150          & 1.391          & 16.17          \\

\textbf{G.SAGE}          & 54.82          & 55.49          & 14.74          & 0.753          & 0.947          & 15.56          \\

\textbf{GAT}                & 53.92          & 56.41          & -5.49          & -0.310         & -0.220         & 24.98          \\

\textbf{SFM}                & 57.02          & 58.30          & 27.76          & 1.269          & 1.461          & 19.00          \\

\textbf{Adv-A.}          & 56.59          & 60.67          & 31.66          & 1.473          & {\underline {2.025}}    & 15.64          \\

\textbf{DTML}               & 54.05          & \textbf{62.44} & 25.37          & 1.215          & 1.560          & 16.27          \\

\textbf{ESTIM.}           & 56.41          & 61.54          & 27.45          & 1.324          & 1.693          & 16.22          \\

\textbf{StockM.}         & 54.01          & 60.00          & 18.39          & 0.914          & 0.982          & 18.72          \\

\textbf{MAST.}             & 57.34          & 59.51          & {\underline {31.70}}    & {\underline {1.517}}    & 2.016          & 15.73          \\

\textbf{GPT4TS}             & 57.01          & 56.09          & 20.23          & 1.047          & 1.320          & {\underline 15.32}    \\

\textbf{aLLM4TS}            & 56.90          & 59.68          & 15.97          & 0.720          & 0.793          & 20.14          \\

\textbf{Time-LLM}           & 56.11          & 56.41          & 24.59          & 1.110          & 1.341          & 18.34          \\

\textbf{H3M-SSMoEs}               & {\underline {57.47}}    & {\underline {62.01}}    & \textbf{50.00} & \textbf{1.585} & \textbf{3.377} & \textbf{14.81} \\
\bottomrule
\end{tabular}
\end{table}

\subsection{Backtesting \& Prediction Results}

We evaluate each model using a dynamic $d$-day trading strategy with adaptive portfolio construction and stop-loss mechanisms, initialized with a capital of $1,000,000$ and assumes a transaction cost of $0.25\%$. Detailed descriptions of the backtesting methodology and hyperparameter configurations are provided in the Appendix \ref{Backtesting Setting}.

\subsubsection{Results for DJIA}

Table \ref{tab:Prediction & Backtesting Results on DJIA} presents the evaluation results on the DJIA. Our model achieves 57.47\% accuracy and 62.01\% precision (second-best after DTML's 62.44\%), demonstrating strong reliability in identifying upward price movements.
In backtesting, H3M-SSMoEs achieves an outstanding annual return of 50.00\%, which is 57.7\% higher than the second-best model, MASTER (31.70\%). 
Risk-adjusted performance metrics further corroborate this superiority, with the highest Sharpe ratio (1.585), the best Calmar ratio (3.377, exceeding the second-best by 66.8\%), and the lowest maximum drawdown (14.81\%). 
These results highlight the efficacy of the H3M-SSMoEs in achieving high returns with controlled risk exposure.

\begin{table}[htb]
\caption{Backtesting \& Prediction Results on NASDAQ 100. The best results are in bold and the second-best results are underlined.}
  \label{Prediction & Backtesting Results on NASDAQ 100}
\begin{tabular}{ccccccc}
\toprule
\multirow{2}{*}{\textbf{Model}} & \multicolumn{2}{c}{\textbf{Prediction}} & \multicolumn{4}{c}{\textbf{Backtesting}}                          \\ \cmidrule(lr){2-3} \cmidrule(lr){4-7}
                                & \textbf{ACC}       & \textbf{PRE}       & \textbf{AR}    & \textbf{SR}    & \textbf{CR}    & \textbf{MDD}   \\
\midrule
\textbf{DLin.}                & 58.52              & 63.50              & 44.82          & 1.491          & 2.737          & {\underline {16.38}}    \\

\textbf{iTrans.}           & 56.28              & 64.94              & 37.83          & 1.447          & 2.151          & 17.59          \\

\textbf{TimeM.}              & {\underline {58.56}}        & 68.42              & 45.56          & 1.602          & 2.259          & 20.16          \\

\textbf{GCN}                    & 56.00              & 58.47              & 19.50          & 0.910          & 0.998          & 19.53          \\

\textbf{G.SAGE}              & 54.87              & 60.68              & 34.92          & 1.217          & 1.826          & 19.12          \\

\textbf{GAT}                    & 53.37              & 59.68              & 32.57          & 1.241          & 1.752          & 18.60          \\

\textbf{SFM}                    & 55.98              & 67.05              & 68.43          & {\underline {2.093}}    & {\underline {4.088}}    & 16.74          \\

\textbf{Adv-A.}              & 56.82              & 63.03              & 55.25          & 1.920          & 3.101          & 17.82          \\

\textbf{DTML}                   & 56.33              & 64.81              & 58.82          & 1.936          & 3.383          & 17.39          \\

\textbf{ESTIM.}               & 53.85              & 51.61              & -7.75          & -0.419         & -0.329         & 23.54          \\

\textbf{StockM.}             & 53.85              & 58.19              & 44.79          & 1.484          & 2.225          & 20.13          \\

\textbf{MAST.}                 & 58.14              & 65.81              & \textbf{71.75} & 1.882          & 3.021          & 23.75          \\

\textbf{GPT4TS}                 & 58.22              & {\underline {69.88}}        & 60.91          & 2.010          & 3.247          & 18.76          \\

\textbf{aLLM4TS}                & 58.25              & 63.70              & 63.93          & 2.085          & 3.682          & 17.36          \\

\textbf{Time-LLM}               & 54.98              & 60.34              & 30.14          & 1.133          & 1.580          & 19.08          \\

\textbf{H3M-SSMoEs}                   & \textbf{58.60}     & \textbf{69.97}     & {\underline {70.80}}    & \textbf{2.100} & \textbf{4.380} & \textbf{16.17} \\
\bottomrule
\end{tabular}
\end{table}

\begin{table}[htb]
\caption{Backtesting \& Prediction Results on S\&P 100. The best results are in bold and the second-best results are underlined.}
  \label{Prediction Backtesting Results on SP 100}
\begin{tabular}{ccccccc}
\toprule
\multirow{2}{*}{\textbf{Model}} & \multicolumn{2}{c}{\textbf{Prediction}} & \multicolumn{4}{c}{\textbf{Backtesting}}                          \\ \cmidrule(lr){2-3} \cmidrule(lr){4-7}
                                & \textbf{ACC}       & \textbf{PRE}       & \textbf{AR}    & \textbf{SR}    & \textbf{CR}    & \textbf{MDD}   \\
\midrule
\textbf{DLin.}                & 56.01              & 60.73              & 25.92          & 1.169          & 1.583          & 16.37          \\

\textbf{iTrans.}           & 55.76              & 64.70              & 27.40          & 1.229          & 1.823          & 15.03          \\

\textbf{TimeM.}              & {\underline {56.74}}        & 50.00              & 19.70          & 0.947          & 1.070          & 18.41          \\

\textbf{GCN}                    & 54.83              & 61.05              & 21.99          & 1.108          & 1.448          & 15.19          \\

\textbf{G.SAGE}              & 55.67              & 61.91              & 24.21          & 1.216          & 1.596          & 15.16          \\

\textbf{GAT}                    & 56.44              & 57.39              & 22.58          & 0.976          & 1.444          & 15.64          \\

\textbf{SFM}                    & 55.65              & {\underline {65.59}}        & 21.76          & 1.080          & 1.404          & 15.50          \\

\textbf{Adv-A.}              & 55.31              & 64.91              & 28.02          & 1.262          & 1.962          & {\underline {14.28}}    \\

\textbf{DTML}                   & 53.80              & 59.22              & 28.20          & 1.305          & 1.869          & 15.09          \\

\textbf{ESTIM.}               & 55.46              & 59.94              & 27.62          & {\underline {1.346}}          & 1.667          & 16.57          \\

\textbf{StockM.}             & 54.51              & 60.43              & 26.71          & 1.335          & 1.859          & 14.37          \\

\textbf{MAST.}                 & 55.17              & 60.59              & 5.08           & 0.246          & 0.325          & 15.63          \\

\textbf{GPT4TS}                 & 55.46              & 60.08              & 27.36          & 1.229          & 1.502          & 18.22          \\

\textbf{aLLM4TS}                & 55.96              & 61.88              & \textbf{30.62}    & {\underline {1.346}}          & {\underline {1.986}}          & 15.41          \\

\textbf{Time-LLM}               & 54.44              & 63.81              & 17.91          & 0.963          & 1.212          & 14.77          \\

\textbf{H3M-SSMoEs}                   & \textbf{56.91}     & \textbf{66.04}     & {\underline {29.62}}          & \textbf{1.351}    & \textbf{2.075}    & \textbf{14.27} \\
\bottomrule
\end{tabular}
\end{table}

\begin{table}[htb]
\small
\setlength{\tabcolsep}{2pt} 
\caption{Ablation results. The best results are in bold and the second-best results are underlined.}
  \label{Ablation results}
\begin{tabular}{cccccccc}
\toprule
\textbf{Dataset} & \textbf{Component} & \textbf{ACC} & \textbf{PRE} & \textbf{AR} & \textbf{SR} & \textbf{CR} & \textbf{MDD} \\ \midrule
\multirow{4}{*}{\textbf{DJIA}}       & \textbf{w/o LCH}   & 57.38          & 53.37          & 16.47          & 0.875          & 1.065          & 15.47          \\
                                     & \textbf{w/o LLM}   & 57.38          & 53.37          & 16.50          & {\ul 0.877}    & 1.067          & 15.47          \\
                                     & \textbf{w/o SSMoEs} & {\ul 57.40}    & {\ul 53.38}    & {\ul 16.52}    & {\ul 0.877}    & {\ul 1.070}    & {\ul 15.43}    \\
                                     & \textbf{H3M-SSMoEs}      & \textbf{57.47} & \textbf{62.01} & \textbf{50.00} & \textbf{1.585} & \textbf{3.377} & \textbf{14.81} \\ \hline
\multirow{4}{*}{\textbf{NASDAQ 100}} & \textbf{w/o LCH}   & 58.12          & 53.16          & 7.40           & 0.345          & 0.331          & 22.36          \\
                                     & \textbf{w/o LLM}   & 57.96          & 52.68          & 9.78           & 0.451          & 0.475          & {\ul 20.60}    \\
                                     & \textbf{w/o SSMoEs} & {\ul 58.18}    & {\ul 52.83}    & {\ul 12.20}    & {\ul 0.535}    & {\ul 0.514}    & 23.73          \\
                                     & \textbf{H3M-SSMoEs}      & \textbf{58.60} & \textbf{69.97} & \textbf{70.80} & \textbf{2.100} & \textbf{4.380} & \textbf{16.17} \\ \hline
\multirow{4}{*}{\textbf{S\&P 100}}   & \textbf{w/o LCH}   & 56.49          & 53.26          & 15.65          & 0.818          & 0.996          & 15.71          \\
                                     & \textbf{w/o LLM}   & 56.54          & 53.27          & {\ul 16.19}    & {\ul 0.845}    & {\ul 1.037}    & 15.62          \\
                                     & \textbf{w/o SSMoEs} & {\ul 56.63}    & {\ul 53.33}    & 16.01          & 0.836          & 1.026          & {\ul 15.61}    \\
                                     & \textbf{H3M-SSMoEs}      & \textbf{56.91} & \textbf{66.04} & \textbf{29.62} & \textbf{1.351} & \textbf{2.075} & \textbf{14.27} \\ \bottomrule
\end{tabular}
\end{table}

\subsubsection{Results for NASDAQ 100}

Table \ref{Prediction & Backtesting Results on NASDAQ 100} presents the evaluation results for the NASDAQ 100 dataset. H3M-SSMoEs achieves the highest accuracy (58.60\%) and precision (69.97\%), indicating superior predictive capability in this highly volatile, technology-driven index.
In backtesting, our model delivers a strong annual return of 70.80\%, second only to MASTER (71.75\%), while demonstrating exceptional risk management. It achieves the best Sharpe ratio (2.100) and the highest Calmar ratio (4.380), significantly surpassing SFM (4.088) and MASTER (3.021). Furthermore, with the lowest maximum drawdown (16.17\%), H3M-SSMoEs exhibits superior ability to generate substantial returns with controlled downside risk.

\subsubsection{Results for S\&P 100}

Table \ref{Prediction Backtesting Results on SP 100} presents the evaluation results on the S\&P 100 dataset. Our model achieves the highest accuracy (56.91\%) and precision (66.04\%), outperforming the second-best TimeMixer (56.74\%) and SFM (65.59\%), respectively, demonstrating robust predictive performance on this diversified index. 
In backtesting, the H3M-SSMoEs delivers competitive returns of 29.62\% (second only to aLLM4TS's 30.62\%) while excelling in risk management. We achieve the best Sharpe ratio (1.351), highest Calmar ratio (2.075, surpassing aLLM4TS's 1.986 by 4.5\%), and lowest maximum drawdown (14.27\%). This combination of near-optimal returns with superior risk-adjusted metrics validates the robustness of our multimodal hypergraph architecture in handling the S\&P 100's diverse constituents.

\subsubsection{Result Visualization}

\begin{figure*}[htbp]
    \centering
    \includegraphics[width=\textwidth]{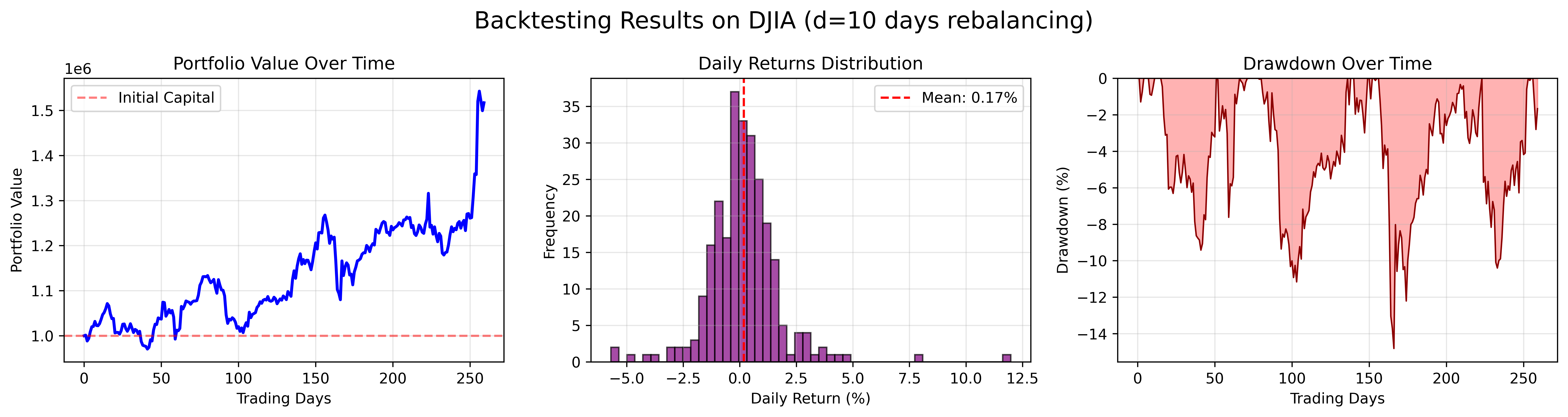}
    \includegraphics[width=\textwidth]{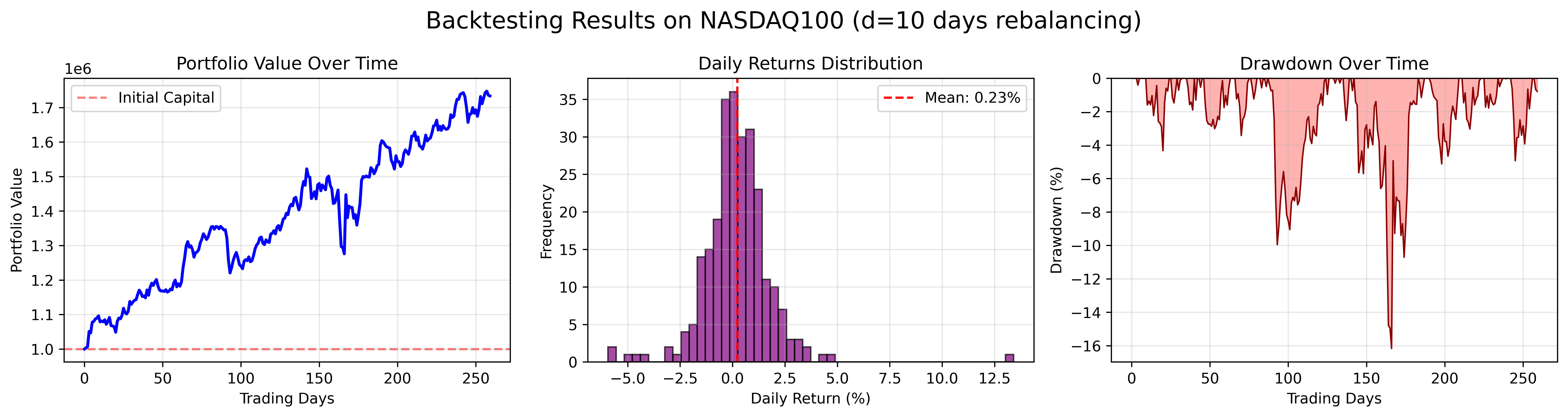}
    \includegraphics[width=\textwidth]{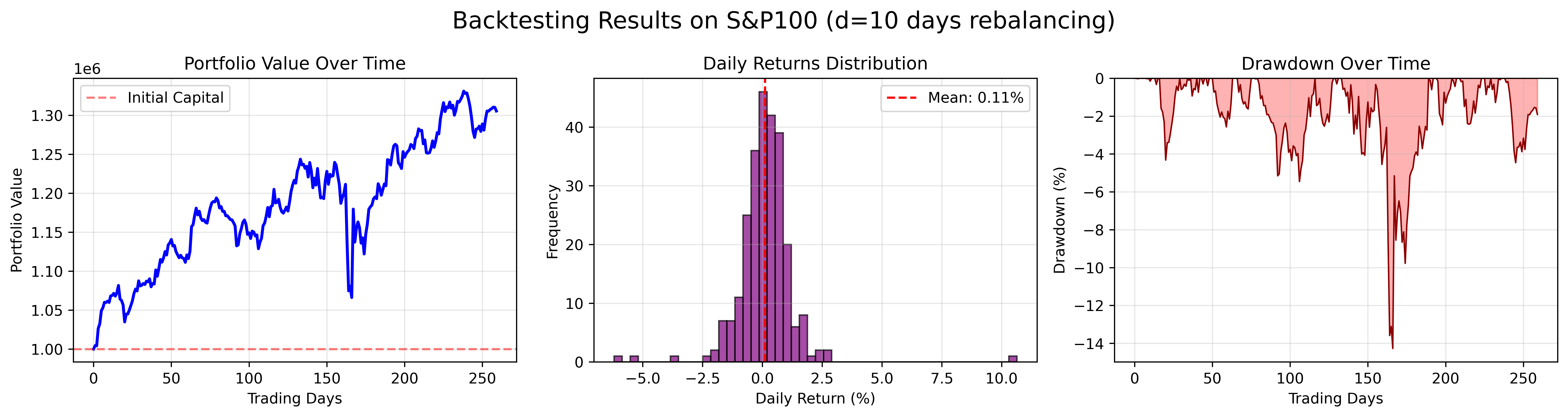}
    \caption{Backtesting performance of the H3M-SSMoEs on DJIA, NASDAQ 100, and S\&P 100 indices
    }
    \label{fig:backtesting_results}
\end{figure*}

Figure \ref{fig:backtesting_results} visualizes  our model's trading performance with 10-day rebalancing periods. Each row corresponds to one index, with three panels illustrating complementary aspects of portfolio behavior. The left panels display portfolio value evolution over testing period, demonstrating consistent upward trajectories. The H3M-SSMoEs achieves the highest terminal value at approximately 1.75x initial capital on NASDAQ 100, followed by DJIA at 1.5x and S\&P 100 at 1.3x. The center panels present daily returns distributions, revealing positively skewed profiles with mean daily returns. The concentration of returns near zero with extended positive tails indicates our model's ability to capture upside opportunities while limiting downside exposure. The right panels depict drawdown dynamics over time, with maximum drawdowns constrained to around 15–17\%, indicating a reasonable risk–return balance. The shaded regions reveal that all three portfolios experience moderate short-term losses, demonstrating the model's resilience and rapid recovery capabilities. Overall, these visualizations corroborate the superior risk-adjusted performance metrics reported in Tables \ref{tab:Prediction & Backtesting Results on DJIA}, \ref{Prediction & Backtesting Results on NASDAQ 100} and \ref{Prediction Backtesting Results on SP 100}, demonstrating that H3M-SSMoEs generalizes well across different market compositions and provides consistent positive returns.

In conclusion, H3M-SSMoEs demonstrates consistent superiority across DJIA, NASDAQ 100, and S\&P 100 indices, achieving the highest Sharpe ratios (1.585, 2.100, 1.351) and Calmar ratios (3.377, 4.380, 2.075) with the lowest maximum drawdowns (14.81\%, 16.17\%, 14.27\%). Combined with competitive or best-in-class returns (50.00\%, 70.80\%, 29.62\%) and high precision (62.01\%, 69.97\%, 66.04\%), these results validate our architectural innovations: multi-context multimodal hypergraph for pattern capture, LLM integration for semantic enhancement, and style-structured MoEs for adaptive specialization. The synergistic combination of these components successfully addresses financial market challenges, achieving superior risk-adjusted returns.

\subsection{Ablation Studies}

To assess the contribution of each architectural component, we conducted ablation studies by removing key modules from H3M-SSMoEs. Table \ref{Ablation results} summarizes the comprehensive results across the three datasets. 
The results demonstrate that each component is indispensable, as its removal leads to substantial performance degradation across all metrics. 

Among the variants, 
removing the Local Context Hypergraph module (\textbf{w/o LCH}) yields the most severe deterioration, with annual returns plummeting from 50.00\% to 16.47\% on DJIA and from 70.80\% to 7.40\% on NASDAQ 100, alongside a reduction in the Calmar Ratio from 4.380 to 0.331 on NASDAQ 100. 
Eliminating the frozen LLM semantic reasoning layer (\textbf{w/o LLM}) produces a similarly adverse effect, reducing annual returns to 9.78\% on NASDAQ 100 (versus 70.80\% for the full model) and decreasing the Sharpe Ratio from 2.100 to 0.451. Likewise, replacing the SSMoEs with a standard feedforward network (\textbf{w/o SSMoEs}) causes notable degradation, with returns declining to 16.52\% (DJIA) and 12.20\% (NASDAQ 100), compared to 50.00\% and 70.80\% achieved by the complete model.

\section{Conclusion}

We proposed the H3M-SSMoEs, a comprehensive multi-modal framework that synergistically integrates multi-context hypergraph modeling, LLM-enhanced semantic reasoning, and a Style-Structured Mixture of Experts (SSMoEs) for stock prediction. By unifying structural, semantic, and stylistic dimensions of market information across quantitative and textual modalities, H3M-SSMoEs effectively captures both fine-grained temporal dependencies and long-term inter-stock relationships while maintaining computational efficiency through lightweight LLM and sparse expert routing. Extensive experiments on DJIA, NASDAQ 100, and S\&P 100 indices show consistent improvements in both predictive accuracy and risk-adjusted returns, achieving state-of-the-art Sharpe and Calmar ratios with significantly reduced drawdowns, validating the robustness and practical applicability of the architecture. Ablation studies further confirm the essential roles of each component. The results demonstrate that integrating hypergraph representations and alignment, LLM reasoning, and adaptive style-structured expert specialization provides a robust foundation for multimodal financial forecasting.

%%
%% The next two lines define the bibliography style to be used, and
%% the bibliography file.
\bibliographystyle{ACM-Reference-Format}
\bibliography{sample-base}

%%
%% If your work has an appendix, this is the place to put it.
\appendix

\section{Problem Definition}
\label{Problem Definition}

We formulate the $d$-day–ahead stock movement prediction task as a binary classification problem. Let $\mathcal{S} = \{s_1, \dots, s_N\}$ denote a universe of $N$ stocks. For each stock $s_i$, we consider its historical quantitative feature over a lookback horizon of $T$ trading days, comprising $F$ financial indicators. These numerical features form a matrix $\mathbf{X}_i^{quant} \in \mathbb{R}^{T \times F}$, and stacking all stock-level matrices yields $\mathbf{X}^{quant} \in \mathbb{R}^{N \times T \times F}$. The ground truth is defined by comparing the closing price on day $t$ with that on day $t+d$:
\begin{equation}
y_i^{(t+d)}=
\begin{cases}
1, & \text{if } p_i^{(t+d)} > p_i^{(t)},\\
0, & \text{otherwise},
\end{cases}\quad \forall i\in{1,\dots,N},
\end{equation}
where $p_i^{(t)}$ denotes the closing price of stock $s_i$ on day $t$.

In addition to quantitative modality, we incorporate two other complementary modalities at each time step:
\begin{itemize}
\item \textbf{Daily news (textual modality)} $\mathbf{X}^{\text{news}} \in \mathbb{R}^{N \times T \times D_{\text{news}}}$: Each stock is paired with one news item per day, which is encoded using a frozen Llama-3.2-1B model \cite{touvron2023llamaopenefficientfoundation}.

\item \textbf{Timestamp embeddings (temporal modality for positional encoding)} $\mathbf{X}^{\text{time}} \in \mathbb{R}^{T \times D_{\text{time}}}$: Each date string is embedded using the same frozen LLM, and the resulting representations are broadcast-added to each stock’s quantitative features at the corresponding time step.
\end{itemize}
Both news and timestamp embeddings are extracted from the LLM’s last hidden state corresponding to the end-of-sequence token $\langle\mathrm{EOS}\rangle$ \cite{liu2024autotimesautoregressivetimeseries}. These embeddings are pre-computed prior to model training to ensure computational efficiency while capturing rich semantic and temporal dependencies.

Let $f(\cdot; \boldsymbol{\Theta})$ denote the predictive function parameterized by $\boldsymbol{\Theta}$. Given the triplet of input sequences $(\mathbf{X}^{quant}, \mathbf{X}^{\text{news}}, \mathbf{X}^{\text{time}})$ over the lookback window, the model estimates the probability of an upward price movement for each stock:
$$
\hat{\mathbf{Y}} = f\left(\mathbf{X}^{quant},\mathbf{X}^{\text{news}},\mathbf{X}^{\text{time}};\boldsymbol{\Theta}\right),
\quad \hat{\mathbf{Y}} \in [0,1]^N.
$$

\section{Global Context Hypergraph (GCH)}
\label{Global Context Hypergraph (GCH)}

Following Local Context Hypergraph (LCH) processing, we transform the features into stock-level representations:
\begin{equation}
    \mathbf{Z}^{(m)}_{flat'} \in \mathbb{R}^{N \times (T \cdot D)}, \quad m \in \{\text{quantitative}, \text{news}\},
\end{equation}
where each row encapsulates the complete temporal evolution of an individual stock. This format facilitates the modeling of long-term, cross-stock relationships that span multiple temporal scales.

Unlike conventional approaches that depend on predefined or static relational structures, financial market interactions are inherently dynamic, continuously forming, evolving, and dissolving as market conditions change. To capture these evolving dependencies, we employ a multi-head attention mechanism that integrates both self- and cross-attention, enabling the model to adaptively learn intra- and inter-modal relationships:
\begin{equation}
\mathbf{A}^{(m_i,m_j)} =
\begin{cases}
\text{MHSA}\left(\mathbf{Z}^{(m_i)}_{flat'}, \mathbf{Z}^{(m_j)}_{flat'}, \mathbf{Z}^{(m_j)}_{flat'}\right), & \text{if } m_i = m_j, \\
\text{MHCA}\left(\mathbf{Z}^{(m_i)}_{flat'}, \mathbf{Z}^{(m_j)}_{flat'}, \mathbf{Z}^{(m_j)}_{flat'}\right), & \text{if } m_i \neq m_j.
\end{cases}
\end{equation}
This mechanism yields four complementary attention matrices, each emphasizing distinct facets of global market dynamics. Analogous to the four sub-hypergraphs in the LCH, these matrices represent quantitative–quantitative interactions, news–news coherence, and bidirectional cross-modal dependencies that connect market behavior with textual narratives.

Financial markets exhibit pronounced collective dynamics, where stocks in the same industry often evolve coherently. To move beyond pairwise attention-based graphs and capture such higher-order interactions, we transform the dyadic attention weights into hypergraph representations:
\begin{equation}
\mathbf{H}^{(m_i,m_j)}_{global} = \text{FFN}^{(m_i,m_j)}_{global}(\mathbf{A}^{(m_i,m_j)}) \in \mathbb{R}^{N \times E_2},
\end{equation}
where $E_2$ denotes the number of global hyperedges, interpretable as latent industry factors. The projection networks $\text{FFN}^{(m_i,m_j)}_{global}$ serve as factorization modules, decomposing dense attention matrices into group membership structures. Each sub-hypergraph is standardized via z-score normalization followed by softmax to ensure valid probabilistic incidence matrices, yielding $\tilde{\mathbf{H}}^{(m_i,m_j)}_{global}$.

The normalized sub-hypergraphs are subsequently integrated through an adaptive fusion network that synthesizes a unified representation:
\begin{equation}
    \mathbf{H}_{global}' = \text{FFN}_{global}^{fusion}([\tilde{\mathbf{H}}^{(m_i,m_j)}_{global}]_{\text{all pairs}}) \in \mathbb{R}^{N \times E_2}
\end{equation}
This fusion module learns non-linear combinations that amplify synergistic relationships while suppressing redundancy. The fused structure is again standardized via z-score normalization and column-wise softmax to produce the final global incidence matrix:
\begin{equation}
    \mathbf{H}_{global} = \mathrm{Softmax}\left( \mathrm{Z\text{-}Score}\left( \mathbf{H}_{global}' \right) \right).
\end{equation}
We also employ JSD-based adaptive weighting to construct a diagonal weight matrix $\mathbf{W_2} \in \mathbb{R}^{E_2 \times E_2}$, which assigns adaptive weights to the hyperedges. Global hypergraph convolutions are then applied to both modalities using the shared incidence matrix $\mathbf{H}_{global}$, enabling consistent propagation of information across all stocks:
\begin{equation}
\mathbf{Z'}^{(m)}_{GCH} = \sigma(\mathbf{H}_{global} \mathbf{W_2} \mathbf{H}_{global}^T \mathbf{X}^{(m)} \boldsymbol{\Theta}_{global}^{(m)}),
\end{equation}
where $m \in \{\text{quantitative}, \text{news}\}$. Unlike the hypergraph convolution in the LCH operating at individual time steps, the GCH convolutions aggregate information across the entire temporal span, thereby capturing persistent patterns in industry behaviors. The resulting features $\mathbf{Z'}^{(m)}_{GCH}$ are reshaped to $\mathbb{R}^{N \times T \times D}$, yielding $\mathbf{Z}^{(m)}_{GCH}$, encapsulates rich, temporally invariant global contextual information.

\section{Expert Routing and Aggregation}
\label{Expert Routing and Aggregation}

For the Shared Market Experts, the routing mechanism determines which experts to activate based on the augmented market representation $\mathbf{z}^{mkt}_i$. The procedure begins by computing routing logits for all market experts:
\begin{equation}
\mathbf{logits}_i^{mkt} = \mathbf{z}^{mkt}_i \cdot\mathbf{W}_{route}^{mkt} + \mathbf{b}_{route}^{mkt} \in \mathbb{R}^{N_{mkt}},
\end{equation}
where $\mathbf{W}_{route} \in \mathbb{R}^{ (D+d_m) \times N_{mkt} }$ denotes the routing matrix that learns to project the concatenated stock feature and market state onto expert relevance scores. To balance computational efficiency with model expressiveness, we employ sparse activation by selecting only the top $K_m$ experts:
\begin{equation}
\text{top\_k\_logits}_i^{mkt}, \text{indices}_i^{mkt} = \text{Top-K}(\mathbf{logits}_i^{mkt}, K_m).
\end{equation}
The sparse gating mechanism subsequently constructs masks, where only the selected experts retain their corresponding activation values while the remainder are masked out:
\begin{equation}
\text{sparse\_logits}_i^{mkt}[j] = \begin{cases}
\text{top\_k\_logits}_i^{mkt}[k], & \text{if } j = \text{indices}_i^{mkt}[k] \\
-\infty, & \text{otherwise}
\end{cases}.
\end{equation}
These sparse logits are normalized via a softmax to yield the gating weights:
\begin{equation}
g_{ij}^{mkt} = \frac{\exp(\text{sparse\_logits}_i^{mkt}[j])}{\sum_{j'=1}^{N_{mkt}} \exp(\text{sparse\_logits}_i^{mkt}[j'])}
\end{equation}
The aggregated market-level output is then computed as a weighted combination of the selected experts’ predictions:
\begin{equation}
\mathbf{h}^{mkt}_i = \sum_{j \in \text{Top-K}(i)} g_{ij}^{mkt} \cdot \text{Expert}_j^{mkt}(\mathbf{z}_i^{flat}),
\end{equation}
where $\text{Top-K}(i)$ represents the subset of $K$ experts activated for stock $i$. This adaptive routing design enables the model to dynamically adjust to evolving market conditions by selectively engaging experts most relevant to the prevailing regime.

In parallel, the industry-specialized experts undergo an analogous routing process, utilizing the industry-augmented representation $\mathbf{z}^{\text{ind}}_i$. The routing logits are computed as:
\begin{equation}
\mathbf{logits}^{ind}_i = \mathbf{z}^{\text{ind}}_i \cdot \mathbf{W}^{\text{ind}}_{route} + \mathbf{b}^{\text{ind}}_{route},
\end{equation}
where $\mathbf{W}^{\text{ind}}_{route} \in \mathbb{R}^{(D+E_2) \times N_{\text{ind}}}$ denotes the industry routing matrix that maps stock features, augmented with learned sectoral embeddings, to industry expert relevance scores. Following the same sparse selection and gating procedure, the aggregated output from industry experts is expressed as:
\begin{equation}
\mathbf{h}_i^{ind} = \sum_{k \in \text{Top-K}(i)} g^{\text{ind}}_{ik} \cdot \text{Expert}^{ind}_k(\mathbf{z}_i^{flat}).
\end{equation}

\section{Experiment Settings}
\label{Experiment Settings}

Our model was implemented in PyTorch and optimized using the cross-entropy loss. Training was conducted for 40 epochs with the AdamW optimizer (learning rate $= 1 \times 10^{-4}$, weight decay $= 0.05$). We applied linear warmup for the first 10\% of training steps, followed by a linear decay schedule. 
The main hyperparameters were configured as follows:
feature embedding dimension $D = 256$,
dropout rate = 0.1,
attention heads = 2 (for GCH),
market state dimension = 16,
expert style dimension = 16,
top-$K = 2$ (for both expert pools in SSMoEs),
auxiliary loss balance factors $\alpha, \beta = 0.1$.
For the LLM backbone, we adopted a frozen Llama-3.2-1B model with a hidden dimension of 2048.
The lookback window was set to $T = 20$ trading days, with a prediction horizon of $d = 10$ days. We tuned several structural hyperparameters through grid search to maximize validation accuracy, including:
number of hyperedges for both LCH and GCH, $E_1, E_2 \in \{32, 64, 128\}$,
number of Shared Market Experts $N_{mkt}\in \{3, 4, 5\}$,
and number of Industry-specialized Experts $N_{ind}\in \{6, 8, 10\}$, and the final settings are reported in Table~\ref{tab:Hyperparameter Configurations}.

\begin{table}[htb]
\caption{Hyperparameter Configurations}
  \label{tab:Hyperparameter Configurations}
\begin{tabular}{cccc}
\toprule
\textbf{Dataset} & $E_1$ \& $E_2$ & $N_{mkt}$ & $N_{ind}$ \\ \midrule
\textbf{DJIA}       & 64 & 3 & 10 \\
\textbf{NASDAQ 100} & 32 & 5 & 6  \\
\textbf{S\&P 100}   & 32 & 3 & 8 \\ \bottomrule
\end{tabular}
\end{table}

\section{Baseline Descriptions}
\label{Baseline Descriptions}
To evaluate the effectiveness of the H3M-SSMoEs, we compare it against 15 baselines with several state-of-the-art baselines from 4 different categories. These models provide a diverse set of benchmarks to evaluate our method’s performance. 

\textbf{1. Stock Prediction Models (6):}
\begin{itemize}
    \item SFM \cite{zhang2017stock}: State Frequency Memory networks that model price fluctuations across multiple frequencies using frequency-based decomposition.
    \item Adv-ALSTM \cite{feng2018enhancing}: Attentive LSTM with adversarial training for improved robustness against stochastic price movements.
    \item DTML \cite{attention1}: Transformer architecture capturing dynamic inter-stock correlations through multi-level contexts.
    \item ESTIMATE \cite{huynh2023efficient}: Combines wavelet-based hypergraph convolution with memory-enhanced LSTM for non-pairwise stock correlations.
    \item StockMixer \cite{fan2024stockmixer}: MLP-based model that sequentially mixes indicators, temporal patterns, and market correlations.
    \item MASTER \cite{li2024master}: Integrates intra/inter-stock attention with market-guided gating for dynamic correlation capture.
\end{itemize}

\textbf{2. Time Series Models (3):}
\begin{itemize}
    \item DLinear \cite{zeng2023transformers}: One-layer linear model that directly models temporal relations for long-term forecasting.
    \item iTransformer \cite{liu2023itransformer}: Inverted Transformer applying attention across variates rather than time steps.
    \item TimeMixer \cite{wang2024timemixer}: MLP-based model using multiscale mixing to disentangle temporal variations.
\end{itemize}

\textbf{3. Graph Models (3):}
\begin{itemize}
    \item GCN (Graph Convolutional Network) \cite{kipf2016semi}: Uses first-order spectral graph convolutions for efficient node embedding learning.
    \item GraphSAGE \cite{hamilton2017inductive}: Inductive framework generating embeddings via neighborhood sampling and aggregation.
    \item GAT (Graph Attention Network) \cite{velivckovic2017graph}: Employs masked self-attention to assign weights to neighbors for flexible node embedding.
\end{itemize}

\textbf{4. Time Series LLM \& Foundation Model (3):}
\begin{itemize}
    \item GPT4TS \cite{zhou2023fitsallpowergeneraltime}: Builds on a frozen GPT-2, fine-tuning only input embeddings, normalization, and output layers, using instance normalization and patching to construct a cross-modality framework for time-series representation.
    \item aLLM4TS \cite{bian2024multipatchpredictionadaptingllms}: Employs a two-stage architecture (causal next-patch pre-training and multi-patch fine-tuning) with a patch-wise decoder to enable localized temporal modeling and representation learning within LLMs.
    \item Time-LLM \cite{jin2024timellmtimeseriesforecasting}: Features a three-part framework consisting of input reprogramming, a frozen LLM backbone, and output projection, where time-series patches are mapped into text prototype embeddings and guided by Prompt-as-Prefix (PaP) prompts for modality alignment.
\end{itemize}

\section{Metric Definitions}
\label{Metric Definitions}

% \subsection{Prediction Metrics}
\begin{equation}
    \text{Accuracy} = \frac{TP + TN}{TP + TN + FP + FN},
\end{equation}
where:
\begin{itemize}
    \item TP (True Positives): Correctly predicted positive cases;
    \item TN (True Negatives): Correctly predicted negative cases;
    \item FP (False Positives): Incorrectly predicted as positive;
    \item FN (False Negatives): Incorrectly predicted as negative.
\end{itemize}

% \subsection{Backtesting Metrics}
\begin{equation}
    \text{Annual Return} = \left[\prod_{t=1}^{T}(1 + r_t)\right]^{\frac{252}{T}} - 1,
\end{equation}
where $r_t$ = return for day t, T = number of trading days, 252 = typical number of trading days per year.

\begin{equation}
    \text{Sharpe Ratio} = \frac{R_p - R_f}{\sigma_p},
\end{equation}
where $R_p$ = annualized portfolio return, $R_f$ = 0.02 (2\% risk-free rate), $\sigma_p$ = annualized standard deviation = $\sigma_{daily} \times \sqrt{252}$.

\begin{equation}
    \text{Calmar Ratio} = \frac{\text{Annual Return}}{|\text{Maximum Drawdown}|},
\end{equation}

\begin{equation}
    \text{Maximum Drawdown} = \min_{t \in [0,T]} \left(\frac{P_t - \max_{s \in [0,t]} P_s}{\max_{s \in [0,t]} P_s}\right),
\end{equation}
where $P_t$ = portfolio value at day t.

\section{Backtesting Setting}
\label{Backtesting Setting}

This section presents a comprehensive description of the dynamic $d$-day stock trading strategy and the corresponding hyperparameter configurations set for backtesting.

Each backtest begins with an initial capital of 1,000,000, incorporating a transaction cost rate of 0.25\% per trade to reflect realistic market frictions. The trading universe comprises all constituent stocks within each index. The core design follows a dynamic $d$-day trading cycle with adaptive portfolio construction and a stop-loss mechanism, where $d$ denotes both the prediction horizon and the rebalancing frequency. The pseudocode for the Dynamic $d$-Day Trading Strategy is presented in Algorithm \ref{alg:d-day-trading-strategy}. All transaction adjustments incorporate transaction costs.

\begin{itemize}
    \item \textbf{Prediction Generation}: The model outputs the probability of price rise $d$ days ahead for all stocks, which we use to rank them.
    \item \textbf{Portfolio Construction with Stop-Loss Mechanism}: We define a portfolio selection proportion $p$ (where $0 < p \leq 1$). On each rebalancing day:
\begin{itemize}
    \item If the number of stocks predicted to rise (with probability $>$ 0.5) $d$ days ahead is at least $p \times N$, we purchase the top $p \times N$ stocks;
    \item If the number of rising predictions falls into an intermediate zone, specifically $p \times N \times q \leq M < p \times N$ (where $q$ is a stop-loss threshold hyperparameter with $0 < q < 1$), then we adopt a conservative approach: only buy the top $r \times M$ predicted rising stocks (with $0 \leq r \leq 1$);
    \item If the number of rising predictions is below $p \times N \times q$, we do not buy new positions and liquidate all current holdings to avoid downside exposure.
\end{itemize}
    \item \textbf{Portfolio Reconstitution}: Positions excluded from the new targets are liquidated with proceeds credited to cash. New target stocks are then purchased with equal capital allocation, subject to current available cash.
    \item \textbf{Portfolio Rebalancing}: To maintain equal-capital allocations, we adjust positions—selling excess holdings that exceed target allocation and purchasing additional shares for under-allocated positions.
    \item \textbf{Hold Period}: Between rebalancing days, all positions are held constant without any trading activity. Portfolio values are recorded daily for performance tracking, but no transactions occur until the next rebalancing day.
\end{itemize}

\begin{algorithm}[htb]
\caption{Dynamic $d$-Day Trading Strategy}
\label{alg:d-day-trading-strategy}
\KwData{$N$ stocks, prediction horizon $d$, Portfolio Selection Ratio $p$, Stop-Loss Thresh-
old $q$, Rising Ratio for Partial Entry $r$, initial capital 1,000,000, transaction cost rate $\tau = 0.25\%$}.
\For{each rebalancing day $t \in \{0, d, 2d, 3d, ...\}$}{
    \tcp{Prediction Generation for d days ahead}
    $\mathbf{P}_t \gets \text{Model.predict\_probabilities}(N \text{ stocks, horizon } d)$\;
    \tcp{$P_{s,t}$ is probability that stock $s$ rises from day $t$ to day $t+d$}
    $M \gets |\{s : P_{s,t} > 0.5\}|$\;
    
    \tcp{Portfolio Construction}
    $n_t \gets \begin{cases}
        \lfloor p \times N \rfloor & \text{if } M \geq p \times N \\
        \lfloor r \times M \rfloor & \text{if } p \times N \times q \leq M < p \times N \\
        0 & \text{if } M < p \times N \times q
    \end{cases}$
    
    \tcp{Portfolio Reconstitution \& Rebalancing}
    \eIf{$n_t = 0$}{
        Liquidate all holdings (apply $\tau$)\;
    }{
        $\text{Targets}_t \gets \text{Top-}n_t\text{ stocks by } \mathbf{P}_t$\;
        Liquidate positions $\notin \text{Targets}_t$ (apply $\tau$)\;
        
        \tcp{Equal capital allocation}
        $\text{TargetValue} \gets \frac{\text{TotalPortfolioValue}}{n_t}$\;
        \For{each stock $s \in \text{Targets}_t$}{
            Adjust position of $s$ to $\text{TargetValue}$ (apply $\tau$)\;
        }
    }
    \tcp{Hold positions constant until next rebalancing day $t+d$}
}
\end{algorithm}

This backtesting strategy facilitates direct evaluation of model predictions and profitability within a realistic trading environment. For our framework, we set $d = 10$ days, balancing prediction reliability with reduced transaction costs from less frequent rebalancing.

\subsection{Backtesting Configurations}
\begin{table}[htb]
\caption{Backtesting Hyperparameter Configurations}
  \label{backtest hyperparameters}
\begin{tabular}{cccc}
\toprule
\textbf{Dataset}    & \textbf{p} & \textbf{q} & \textbf{r} \\ \midrule
\textbf{DJIA}       & 1                & 0.05                 & 0.05                    \\
\textbf{NASDAQ 100} & 1                & 0.05                 & 0.15                    \\
\textbf{S\&P 100}   & 1                & 0.65                 & 0.25                  \\ \bottomrule
\end{tabular}
\end{table}
For each model–dataset pair, we perform grid search on the validation set across three hyperparameters in the trading strategy:
Portfolio Selection Ratio $p \in \{0.05, 0.10, \ldots, 1.0\}$,
Stop-Loss Threshold $q \in \{0.05, 0.10, \ldots, 0.95\}$, and
Rising Ratio for Partial Entry $r \in \{0.0, 0.05, \ldots, 1.0\}$.
The optimal combination yielding the highest Sharpe ratio on the validation set is applied to the test set for final evaluation. The selected hyperparameters for each dataset of our model are shown in Table \ref{backtest hyperparameters}.

\end{document}